# Human-AI Teaming Co-Learning in Military Operations

Clara Maathuis*[a], Kasper Cools[b,c]
[a]Open University of the Netherlands
[b]Belgian Royal Military Academy
[c]Vrije Universiteit Brussel

**ABSTRACT**

In a time of rapidly evolving military threats and increasingly complex operational environments, the integration of AI into military operations proves significant advantages. At the same time, this implies various challenges and risks regarding building and deploying human-AI teaming systems in an effective and ethical manner. Currently, understanding and coping with them are often tackled from an external perspective considering the human-AI teaming system as a collective agent. Nevertheless, zooming into the dynamics involved inside the system assures dealing with a broader palette of relevant multidimensional responsibility, safety, and robustness aspects. To this end, this research proposes the design of a trustworthy co-learning model for human-AI teaming in military operations that encompasses a continuous and bidirectional exchange of insights between the human and AI agents as they jointly adapt to evolving battlefield conditions. It does that by integrating four dimensions. First, adjustable autonomy for dynamically calibrating the autonomy levels of agents depending on aspects like mission state, system confidence, and environmental uncertainty. Second, multi-layered control which accounts continuous oversight, monitoring of activities, and accountability. Third, bidirectional feedback with explicit and implicit feedback loops between the agents to assure a proper communication of reasoning, uncertainties, and learned adaptations that each of the agents has. And fourth, collaborative decision-making which implies the generation, evaluation, and proposal of decisions associated with confidence levels and rationale behind them. The model proposed is accompanied by concrete exemplifications and recommendations that contribute to further developing responsible and trustworthy human-AI teaming systems in military operations.

## 1. INTRODUCTION

Modern defence organisations are confronted with accelerated threat developments and adversaries that blend cyber, electromagnetic, and kinetic actions with unprecedented fluidity [1], [2]. Against such threats, AI (Artificial Intelligence)-enabled systems now drive applications [3] such as sensor fusion, threat classification, and Courses of Action (CoAs) generation with gains in both tempo and precision as well as resilience. Nevertheless, building and deploying AI models into C2 (Command and Control) chains without dedicated responsible measures can introduce various new failure or intentional sensitive nodes such as brittle models, opaque reasoning, mis-calibrated trust, and accountability gaps that jeopardise compliance with the law of armed conflict [4]. Therefore, human-AI teaming demands a design paradigm that treats the agent pair not as a monolithic super-agent, but as a dynamic socio-technical system whose internal feedback, workload shifts, and authority transfers critically shape combat effectiveness and legal and ethical compliance [5], [6]. Specifically, human-AI teaming represents the partnership between human and AI agents where their complementary strengths are combined to achieve shared goals [7], [8] while embedding human qualities like ethical judgement, adaptability, and creativity alongside AI's capabilities in data processing, pattern recognition, and automation [9], [10]. The core roles in a human-AI teaming setting include dynamic coordination, mutual adaptation, and the development of shared mental models, where the teammate agents are distinct team members who adjust own behaviours in response to changing tasks and activities in a specific context.

Human-AI teaming systems are built in applications like cancer detection where radiologists work alongside AI agents to assure and improve the accuracy and efficiency of the diagnosis [11]. In the creative domain, digital artists employ AI tools to generate novel design concepts and streamline the creative process[12]. In manufacturing, workers team with collaborative robots (cobots) to optimize production workflows and enhance safety [13].

*Corresponding author: Maathuis C., e-mail: clara.maathuis@ou.nl

At the same time, in the military domain, human operators and AI-driven decision support systems jointly analyse battlefield data to inform tactical decisions, while unmanned aerial vehicles (UAVs) equipped with AI collaborate with human commanders to conduct surveillance, reconnaissance, and coordinated mission planning [14], [15], [16]. The learning component between the human agent and the AI agent is a pillar for effective human-AI teaming, especially in critical domains such as the military domain as this function allows the agents to continuously learn from each other and adapt their behaviours which is essential for maintaining operational effectiveness, safety, and mission success in the face of rapidly changing and unpredictable circumstances [17]. Co-learning processes facilitate the development of shared mental models, mutual trust, and calibrated reliance, ensuring that both agents understand not only the task environment but also each other's capabilities, limitations, and decision-making patterns [18]. This dynamic adaptation is realized through mechanisms such as joint training exercises, real-time feedback loops, and the integration of human experiential knowledge with AI's data-driven insights. Despite growing interest in human-AI teaming, there remains a knowledge gap in the scientific literature and practitioner efforts regarding models that explicitly capture and represent the co-learning processes between human and AI agents. Therefore, this research aims to develop a model that captures and represents the key elements of co-learning in human-AI teams, with a particular focus on military operations. By addressing this gap, the proposed model seeks to advance understanding of how ongoing learning and adaptation between human and AI agents can be executed to enhance team performance, trust calibration, and mission success. On this behalf, the model is developed using the System Dynamics mechanism [19] following the Design Science Research methodology [20].

The model embeds four characteristics that are dynamic autonomy calibration, multi-layer oversight, bidirectional explanation exchange, and confidence-scored collaborative decision-making in order to assure that AI is not perceived as a replacement but as an adaptive teammate whose competence, confidence, and constraints evolve with its human counterpart when planning, executing, and assessing military operations. Specifically, dynamic autonomy calibration ensures that the level of autonomy of the AI agent is continuously adjusted in response to evolving mission requirements and the preferences of the human agent, thereby supporting flexible and context-sensitive teaming. Further, multi-layer oversight enables both human and AI agents to monitor, guide, and intervene in each other's actions at various levels, promoting transparency, accountability and operational safety. At the same time, bidirectional explanation exchange fosters mutual understanding by enabling both agents to articulate their reasoning and intentions, aspects that are essential for building trust and shared situational awareness among them and at team level. Finally, collaborative decision-making integrates assessments of both human and AI agents confidence into joint decisions, enhancing transparency and reducing the risk of over- or under-reliance on the recommendations proposed by the AI agents. To this end, the model is demonstrated for evaluation purposes on a use case conducted and implemented surrounding the execution of the proportionality assessment in military operations. Based on the results obtained, the model proves to be effective in this context and represents a modelling and simulation basis for further understanding and enhancing human-AI teaming efforts and systems in this domain.

The outline of this article is structured as follows. In Section 2, related research studies are discussed. In Section 3, the research methodology followed in this research is presented. In Section 4, the design of the System Dynamics model proposed is tackled. In Section 5, the simulation settings and results obtained are discussed. In Section 6, concluding remarks and future research perspectives are presented.

## 2. RELATED RESEARCH

Human-AI teaming has emerged as a critical paradigm across high-stakes domains, with civilian applications demonstrating improved outcomes in healthcare diagnostics [11] and manufacturing workflows [13]. In military contexts, research has primarily focused on AI-driven decision support for tactical analysis and battlefield data interpretation [14], [16]. Recent US Army systems engineering workshops have similarly identified human-machine co-learning as a critical long-term research outcome, emphasizing the need for new methodological frameworks in military AI applications [21]. However, these studies often adopt an external systems perspective, treating human-AI teams as collective agents rather than examining the dynamics needed for effective collaboration.

The deployment of AI in military operations raises significant concerns regarding trustworthiness and accountability. [4] highlights the risks of integrating AI into C2 chains without sufficient safeguards, including brittle models and mis-calibrated trust that may compromise adherence to international law. [22] further demonstrates that trust calibration varies across military applications (data integration, autonomous systems, and decision-support) with severe consequences when misaligned. Despite widespread recognition of trust's importance, existing mitigation strategies remain insufficient for dynamic combat environments.

Additionally, co-learning between human and AI agents is increasingly recognized as essential for operational effectiveness [17]. [18] identify six key challenges for adaptive human-AI collaboration, emphasizing the need for AI agents to develop mental models with self-awareness and theory of mind capabilities to effectively communicate their internal states. Recent empirical work by [23] implemented Learning Design Patterns in a human-robot urban search and rescue testbed, demonstrating that structured co-learning interventions can enhance human understanding and partner modelling - though improved understanding did not necessarily translate to better team performance. [24], examining soldiers' perspectives in route clearance operations, further underscore the importance of bidirectional communication and mutual understanding. Nevertheless, explicit models capturing co-learning processes remain underdeveloped, with current approaches often focusing on individual agent learning rather than reciprocal exchange.

Autonomy management presents another critical challenge. [5] advocate for viewing human-AI teams as dynamic socio-technical systems, where authority transfers significantly influence team effectiveness. Yet, existing frameworks typically rely on static autonomy levels that fail to adapt to evolving operational conditions, thereby limiting their effectiveness in dynamic military environments [1], [2]. Despite these advances, a cohesive framework that integrates adjustable autonomy, bidirectional feedback, and collaborative decision-making remains underdeveloped. This paper addresses this gap by proposing a co-learning model tailored to dynamic military operations.

## 3. RESEARCH METHODOLOGY

This research follows the Design Science Research [20] methodology in order to design and evaluate the trustworthy co-learning model applicable in human-AI teaming settings in the planning, execution, and assessment of military operations. To this end, a structured extensive literature review is performed covering academic and practitioner military literature and studies carried out in the AI, human-AI teaming, responsible and trustworthy AI, and military domains. This implies the formulating and executing of search queries in scientific databases like IEEE Xplore, ACM Digital Library, Springer, Wiley, and NATO STO and CCDCOE archives from the last five years using combinations of keywords such as human-AI teaming, human-autonomy teaming, responsible AI, trustworthy AI, AI learning, military operations, military targeting, and proportionality. From here, the design requirements and potential modelling and simulation techniques are considered. Taking into consideration the human-AI teaming context and the co-learning dynamics and characteristics, the artefact proposed in this research is a System Dynamics (SD) model. SD is a modelling and simulation mechanism that allows building understanding, analyse, and simulate complex systems that are characterized by interdependencies, feedback loops, time delays, and nonlinear relationships [25]. As explained in the next section, this mechanism uses flow diagrams, stocks, and differential equations to represent how various components of a system interact and evolve over time.

Guided by these requirements, the model is built in an iterative way in Python. In the development phase, four core dimensions were formalised: adjustable autonomy, bidirectional knowledge transfer, multi-layered control, and collaborative decision-making. In order to demonstrate and evaluate the model proposed, a use case focusses on the execution of the proportionality assessment carried out when planning military operations. The results obtained effectively depict the various dynamics that characterize this process and reflect the effectiveness of considering such a modelling and simulation mechanism in this context.

## 4. MODEL DESIGN

SD is a modelling and simulation mechanism designed to understand and simulate the behaviour of complex systems over time. This provides a formal agentic framework for representing systems using stocks, flows, feedback loops, and time moments in order to be analysed in relation to their structure, interconnections, and interdependencies within to capture their dynamics. This mechanism allows the exploration of system evolution considering specific context factors and allows the integration of feedback loops within [19], [26].

As this research proposes a SD model for capturing and representing co-learning in human-AI teaming settings in military operations, it considers this collaborative agent as being a sociotechnical system characterized by intricate feedback loops, mutual adaptation, and evolving trust between human and AI agents. The SD approach is suitable on this behalf as it allows for the explicit modelling of feedback mechanisms such as how changes in trust, shared situation awareness, or workload influence subsequent team behaviour and learning outcomes over time. Accordingly, the design of the model is further discussed.

The stocks are the resources and knowledge that capture the essential characteristics and values of the system analysed that have a static nature. In this research they are defined as follows:

- Human expertise $H$ which represents the mission specific knowledge of the human agent.
- AI competence $A$ which captures the accuracy and generalization ability of the AI agent in the current context.
- Shared situation awareness $S$ which is the alignment of the mental models of teammate agents about the mission state.
- Trust calibration $T$ which represents the trust calibration of the human agent in the AI agent.
- AI authority level $U$ which points out to the degree of adjustable autonomy of the AI agent.
- Cognitive load $C$, which is the mental workload, experienced by the human agent.

The flows are the dynamic component of the system and contain the rates at which stocks increase or decrease over time based on transferred information. In the SD model, the dynamic behaviour of the system emerges from interaction between these stocks and the flows that increase or decrease their levels, allowing for the analysis of how complex systems develop and respond to interventions over time. The flows are defined as follows:

- Human learning $\dot{H}$ which contains the knowledge and feedback from the AI agent that the human agent has minus the cognitive load of the human agent:
$$\dot{H} = \propto_1 F_{HA} - \beta_1 C$$
- AI learning $\dot{A}$ which contains the knowledge and feedback from the human agent that the AI agent has minus the decay term that represents the concept drift present due to issues such as noise and environmental volatility:
$$\dot{A} = \propto_2 F_{AH} - \beta_2 \sigma_{env}$$
- Situation awareness alignment $\dot{S}$ which captures how closely the human agent and the AI agent share the same mental model of the context, i.e., the awareness fusion between the two agents through means such as updates and intent visualizations minus the gap that an agent has:
$$\dot{S} = \gamma_1 F_{sync} - \gamma_2 \Delta_{obs}$$
- Trust calibration $\dot{T}$ which represents trust as a performance sensitive and explanation governance state, where the first component shows the comparison between the competence of the AI agent against the human agent, and the second component prevents the reliance on opaque or poorly justified recommendations made by the AI agent:
$$\dot{T} = \delta_1 PG - \delta_2 EQ$$
- Authority adjustment $\dot{U}$ which represents the delegated authority composed by the two components that capture the calibrated trust and shared situation awareness as positive enablers minus the risk control/brake that automatically retracts autonomy when the operating environment becomes volatile:
$$\dot{U} = k_1 T + k_2 S - k_3 \sigma_{env}$$
- Cognitive load $\dot{C}$ which shows how the mental workload evolves as mission tempo and delegation fluctuate. Specifically, the first component is the task rate of conversion between external elements like sensor or fire-mission updates into incremental workload. And the second component is the off-loading effect of autonomy where the element in the parenthesizes shows the fraction of work still handled manually:
$$\dot{C} = \theta_1 TR - \theta_2 (1 - U)$$

Furthermore, the feedback structures contain the following elements:

- Co-learning reinforcing loop *R1* which shows that the better the explanations from the AI agent are, the faster the learning will take place for both the human and team agents.
- Adjustable autonomy balancing loop *B1* which points out to the fact that when the mission uncertainty $\sigma_{env}$ grows, the team agent lowers the authority *U.*
- Multi-layered control loop *B2* which refers to the situation when *U* increases, the supervisory load increases as well which brings further additional accountability reviews from the human agent and a demand for further explanations from the AI agent.
- Trust performance reinforcing loop *R2* which refers to the situation when the performance of the AI agent exceeds expectations which can also conduct to a raise of *U*, the use of more or additional data which may further imply a performance improvement and a higher computational cost.
- Cognitive load safety loop *B3* which shows that when *C* is raised above a safety threshold automatically will decrease *U* and supress *TR* in order to protect the overload on the human agent and unsound or invalid decisions.

Hence, the four components of the model proposed are described in Table 1 below.

| No. | Dimension | SD Element | Key loops |
|---|---|---|---|
| 1 | Adjustable autonomy | $U$ and $\mathring{U}$ | B1 and B3 |
| 2 | Multi-layered control | Oversight pressure and accountability delays | B2 |
| 3 | Bidirectional feedback | $F_{HA}, F_{AH},$ and $F_{sync}$ | R1 and R2 |
| 4 | Collaborative decision-making | Decision quality depending on *H, A, S,* and *C* | R1, R2, and B3 |

Table 1. Model components

## 5. USE CASE

Following the principles of the DSR methodology, for demonstration and evaluation purposes, a use case is built for the proposed model focusing on the execution of the proportionality assessment integral to lawful military targeting. In this assessment, the expected collateral damage on civilians and civilian objects are weighted against the anticipated military advantage that would contribute to the achievement of military objectives in an operation is conducted [27], [28].

In a hypothetical joint force operation, a cyber-kinetic strike is planned for neutralising an adversary's Integrated Air-Defence Network (IADN) in a coastal city that represents an urban hub with infrastructure is shared by both military and civilian services. The plan centres on a remote zero-day exploit that will disable the IADN's command-and-control servers for 60 minutes, creating a window for manned Suppression of Enemy Air Defence (SEAD) aircraft to penetrate contested airspace. In the mission-planning cell, a human operator (*H* = 0.50) and an AI decision-support agent (*A* = 0.40) jointly perform the proportionality assessment. Environmental uncertainty is elevated ($\sigma_{env}$ = 0.35) because collateral disruption could sever hospital telemetry links and maritime traffic controls on the civilian spectrum. The AI furnishes moderately rich counterfactual graphs and confidence bounds (explanation_quality = 0.75), while the operator labels potential collateral nodes and ethical constraints (annotation_quality = 0.65). Shared situation awareness is initially limited (*S* = 0.25), trust cautious (*T* = 0.40), delegated autonomy low (*U* = 0.20), and cognitive load moderate (*C* = 0.30). Furthermore, the oversight policy stipulates strict governance: authority may rise only as trust and situational awareness converge ($k_1$ = 0.45, $k_2$ = 0.35) and must fall whenever uncertainty spikes ($k_3$ = 0.25). Throughout the one-hour planning window the team iteratively evaluates the anticipated military advantage, i.e., suppressing the IADN without kinetic fire, against the expected collateral damage, i.e., civilian fibre outages that exceed legal thresholds.

The resulting simulation captured interactions among human expertise, AI competence, shared situation awareness, trust, delegated authority, and cognitive load, producing a time-resolved proportionality assessment that balances mission advantage against collateral damage risk. Accordingly, the simulation results are captured in Figure 1 and Figure 2 below.

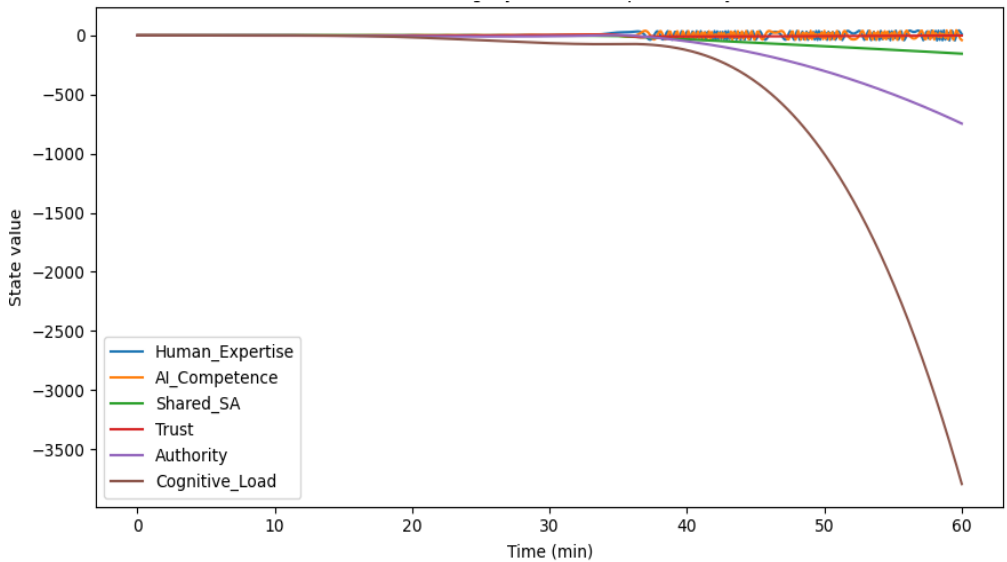

Figure 1. Use case simulation for proportionality assessment

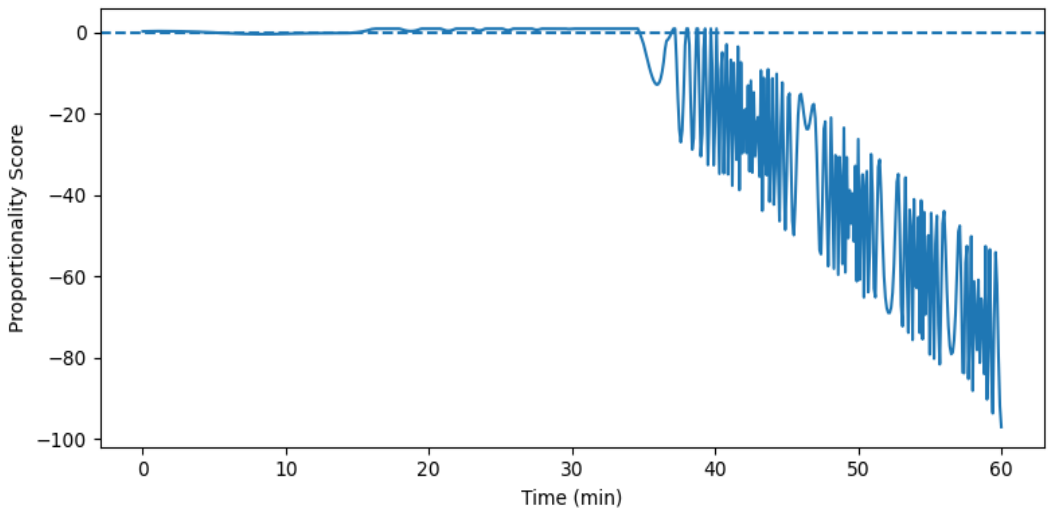

Figure 2. Use case simulation for proportionality assessment

The simulation that the co-learning architecture can initially generate a lawful balance between military advantage and collateral-damage risk, but that this equilibrium is fragile under high environmental uncertainty. During the first 30 minutes, bidirectional feedback raises both human expertise and AI competence, synchronising mental models (shared situation awareness) and calibrating trust. This results in delegated authority which rises cautiously, peak cognitive load that remains below overload thresholds, and the proportionality score stays marginally positive. Nevertheless, when environmental volatility exists ($i.e., \sigma_{env}$ = 0.35) it implies dense civilian presence and disruption which is reflected in the trust collapse faster than the oversight loop which can taper AI autonomy, triggering a precipitous withdrawal of authority. Because military advantage in the model is partly authority-driven, it falls sharply into negative territory while collateral damage, although capped by the autonomy clamp, stays elevated. The net effect is a proportionality score that is positive for only 44 % of the mission window, failing the legal sufficiency criterion and forcing a delay execution until safeguards are taken, abort, or re-plan decision to retain mission effectiveness and be legally compliant.

Hence, the results show the effectivity and usefulness of the model proposed: it reproduces expected doctrinal behaviour (autonomy retraction under uncertainty, workload-driven safety loops) and quantifies the trade-off between aggressive delegation and legal robustness. Future iterations should couple $\sigma_{env}$ to a real-time risk feed and embed a stochastic collateral-damage estimator, enabling commanders to test whether enhanced explainability or adaptive autonomy policies can sustain a positive proportionality score across a broader envelope of urban and hybrid-warfare context conditions.

## 6. CONCLUSIONS

This research presents a trustworthy co-learning model for human-AI teaming in military operations, addressing critical gaps in current approaches to AI integration within defence systems. Developed using System Dynamics methodology within a Design Science Research framework, the four-dimensional model—encompassing adjustable autonomy, multi-layered control, bidirectional feedback, and collaborative decision-making—offers a comprehensive foundation for understanding and optimizing human-AI partnerships in high-stakes military environments. The proportionality assessment use case demonstrates the model's practical applicability. It reveals that while the co-learning architecture can generate lawful balances between military advantage and collateral damage risk through bidirectional feedback and trust calibration, this equilibrium remains fragile under high environmental uncertainty. Simulation results reflect expected doctrinal behaviour, such as autonomy retraction under uncertainty and workload-driven safety loops, while quantifying critical trade-offs between aggressive delegation and legal robustness in military decision-making. The proposed framework enables systematic exploration of how interventions—such as new training protocols, interface modifications, or AI upgrades—propagate through the team, supporting both theoretical insights and operational decision-making. Future research should focus on coupling environmental uncertainty to real-time risk feeds, embedding stochastic collateral damage estimators, and validating the model across diverse military contexts to enhance generalizability. As military organizations continue to integrate AI capabilities, this co-learning model provides a foundation for developing systems that truly partner with human operators—adapting and learning together while upholding ethical standards and preserving the accountability essential for responsible military operations in an era of rapidly evolving threats and operational complexity